%% file: main_final_v4.tex
\documentclass[letterpaper, 10 pt, conference]{ieeeconf}  

\useRomanappendicesfalse

\usepackage{amsmath,amsfonts}
\usepackage{algorithm}
\usepackage{amssymb,algpseudocode}
\usepackage{array}
\usepackage[caption=false,font=normalsize,labelfont=sf,textfont=sf]{subfig}
\usepackage{textcomp}
\usepackage{stfloats}
\usepackage{url}
\usepackage{verbatim}
\usepackage{graphicx}
\usepackage{cite}
\input{math_commands.tex}

\usepackage{xcolor}
\usepackage{soul}

\IEEEoverridecommandlockouts                              

\title{\LARGE \bf
{Uniform Transformation}:\\ Refining {Latent Representation} in {Variational Autoencoders}
}

\author{Ye~Shi
and~C.~S.~George~Lee
\thanks{Ye Shi and C.S. George Lee 
are with the Elmore Family School of Electrical and Computer Engineering, Purdue University, West Lafayette, IN 47907, USA. 
Emails: {\tt\small \{shi349, csglee\}@purdue.edu}
}
}

\begin{document}

\maketitle
\thispagestyle{plain}
\pagestyle{plain}

\begin{abstract}
Irregular distribution in latent space causes posterior collapse, misalignment between posterior and prior, and ill-sampling problem in Variational Autoencoders (VAEs).
In this paper, we introduce a novel adaptable three-stage Uniform Transformation (UT) module -- Gaussian Kernel Density Estimation (G-KDE) clustering, non-parametric Gaussian Mixture (GM) Modeling, and Probability Integral Transform (PIT) -- to address irregular latent distributions. 
By reconfiguring irregular distributions into a uniform distribution in the latent space, our approach significantly enhances the disentanglement and interpretability of latent representations, overcoming the limitation of traditional VAE models in capturing complex data structures. 
Empirical evaluations demonstrated the efficacy of our proposed UT module in improving disentanglement metrics across benchmark datasets -- dSprites and MNIST. 
Our findings suggest a promising direction for advancing representation learning techniques, with implication for future research in extending this framework to more sophisticated datasets and downstream tasks.
\end{abstract}

\section{Introduction}\label{sec:introduction}
Latent representations, crucial for transferring and sharing significant heuristics across tasks, have garnered increased attention due to advancements in optimization~\cite{gonzalez_solving_2022}, foundation models and artificial general intelligence~\cite{gaoSurveyFoundationModels2024}. 
A common approach to obtaining these representations is through variational autoencoders (VAEs)~\cite{kingmaAutoEncodingVariationalBayes2013}, which compress high-dimensional data into a compact, low-dimensional, and disentangled latent space.

\begin{figure}[htb]
    \centering    
    \includegraphics[width=\columnwidth]{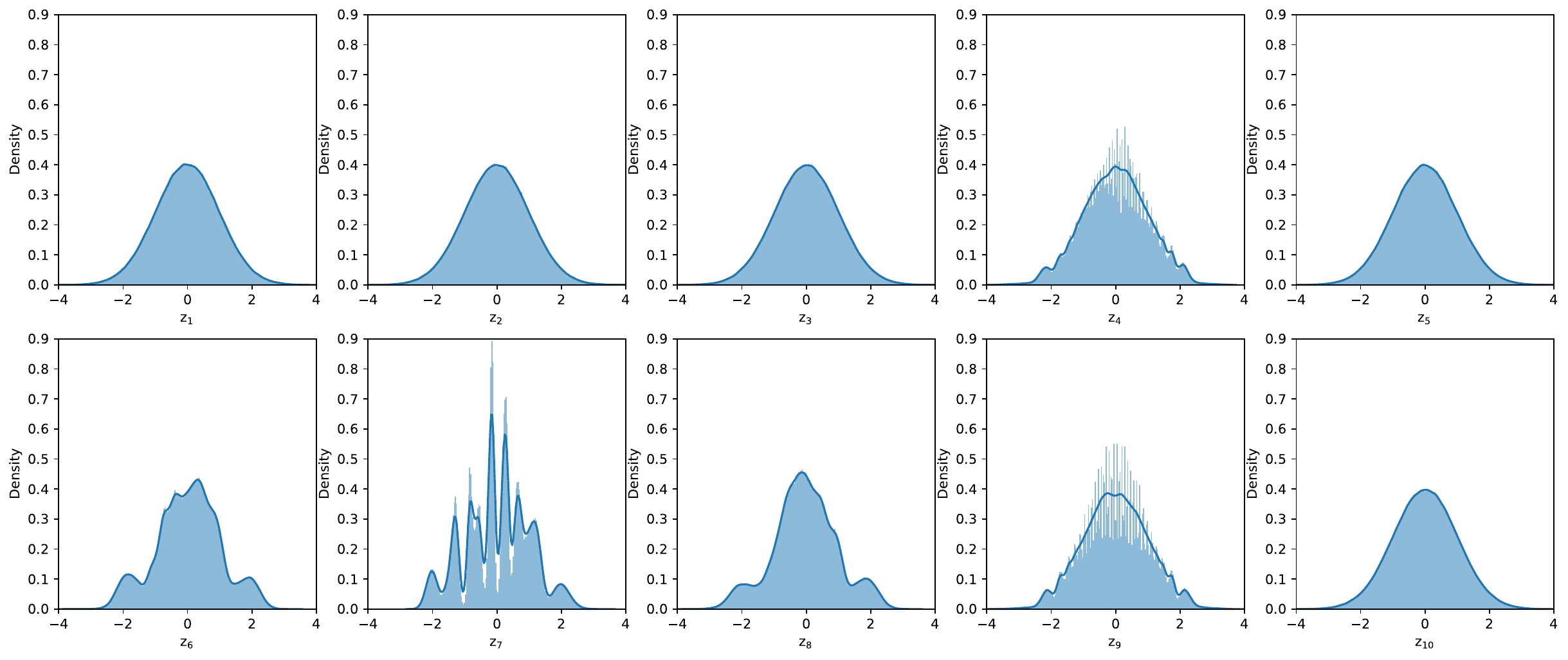}
    \caption{Learned Latent Variable Histograms with Irregular Distributions from Replicating the $\beta$-VAE of dSprites Dataset~\cite{higginsBetaVAELearningBasic2016a}.
    $\rz_7$ encodes the shape and scale labels together. 
    $\rz_6$ and $\rz_8$ encode the rotation label.
    The encoding correspondence is analyzed by correlation in Appendix~\ref{apx:betavae}.
    }
    \label{fig:betaVAE_dist}
\end{figure}

However, VAEs often introduce irregular posterior distributions due to their nonlinear compression nature~\cite{khemakhemVariationalAutoencodersNonlinear2020b}, altering the original data distribution and potentially leading to inefficiency in representation~\cite{nazabalHandlingIncompleteHeterogeneous2020,koike-akinoAutoVAEMismatchedVariational2022}.
Specifically, encoding disentangled categorical features without altering their structure remains a challenge~\cite{dupontLearningDisentangledJoint2018,jeongLearningDiscreteContinuous2019, nazabalHandlingIncompleteHeterogeneous2020},
as these features often converge into singular latent variables with irregular distributions. 
These latent variables typically distribute in an irregular mixture fashion as depicted in~\Figr{fig:betaVAE_dist}.
While early research~\cite{burdaImportanceWeightedAutoencoders2016} viewed this as undesirable due to potential decoder collapse,
recent studies~\cite{aspertiSurveyVariationalAutoencoders2021} highlight it as a sparsity property that balances between representation learning and generative performance. 
Although these latent variables are uncorrelated, the data features remain entangled within singular latent variables~\cite{chenIsolatingSourcesDisentanglement2019}. 
This process disrupts the continuity in the latent representation. 
When sampling from the latent space results in discontinuous posteriors, it leads to challenges in uncertainty quantification~\cite{chandraRevisitingBayesianAutoencoders2022}.

Another challenge arises with periodic data distributions and their sampling~\cite{naimanGenerativeModelingRegular2023}. 
VAEs assume that latent representations are independent and identically distributed in isotropic Gaussians, 
an assumption not always suitable for periodic data~\cite{koike-akinoAutoVAEMismatchedVariational2022}. 
For example, $\rz_6$ and $\rz_8$ in~\Figr{fig:betaVAE_dist} 
encode a continuous rotation label. 
However, shapes with central symmetry, such as a square, can result in indistinguishable data representations when rotated by 90 degrees.
Hence, periodic data features and ill-sampled data are usually poorly encoded in VAEs.
These irregular distributions, which impact the accuracy and utility of the learned representations,
stem from the imbalance of data features in a balanced latent space~\cite{nazabalHandlingIncompleteHeterogeneous2020}. 
This results in the sampling of invalid optima among under-represented data features during the categorical optimization process.
Understanding and addressing these limitations is crucial for improving effectiveness in complex data representation tasks.

In response to these challenges, researchers have focused on regulating latent representations in VAEs using Gaussian Mixture (GM) distributions across various tasks~\cite{dilokthanakulDeepUnsupervisedClustering2017a, suDetectingOutlierMachine2022}.
Often, they might overlook that irregular VAE distributions can be effectively modeled as pre-designed GMs.
Efforts have involved adversarial structures with external discriminators (e.g., AAE\cite{berthelotUnderstandingImprovingInterpolation2018}, FactorVAE~\cite{kimDisentanglingFactorising2018}) or deriving analytical loss functions (e.g., GM-VAE~\cite{dilokthanakulDeepUnsupervisedClustering2017a}, TCVAE~\cite{chenIsolatingSourcesDisentanglement2019}, RAE~\cite{ghoshVariationalDeterministicAutoencoders2020}) to align latent distributions with GMs, highlighting the versatility of VAEs in capturing complex data. 
Some approaches discretize the latent space (e.g., VQ-VAE~\cite{vandenoordNeuralDiscreteRepresentation2017}, Joint-VAE~\cite{dupontLearningDisentangledJoint2018}, CascadeVAE~\cite{jeongLearningDiscreteContinuous2019}) by designing discrete latent dimensions. 
However, this relies on inductive bias, questioning whether the disentanglement arises from prior knowledge or unsupervised learning. 
Coincidentally, these GM regulations often assume balanced child distributions, enumerating a discrete uniform distribution. 
Uniform distributions in VAEs enhance generative capabilities, enable flexible posterior-prior pairings, and improve latent interpretation~\cite{imCausalEffectVariational2022,koike-akinoAutoVAEMismatchedVariational2022}.

To further explore this problem, we analyze it by formulating it as a constrained optimization problem, explaining how irregular distributions are formalized. 
We discovered that these distributions are resulted from a combination of the hard equality constraint from the reparameterization trick forming each child-Gaussian distribution
and the soft constraint of the Kullback–Leibler (KL) divergence that bounds the dimension-wise distribution.
The irregular latent distribution is identified as a GM distribution with a normal distribution constraint.
It also guides us in estimating the latent distributions, a challenging task in distribution transformation.

In this paper, we propose a three-stage uniform transformation (UT) module that transforms GM-distributed latent space to uniform-distributed latent space.
In Stage 1, upon identifying the irregular distribution is a GM distribution, 
we propose a non-parametric univariate clustering algorithm based on Gaussian Kernel Density Estimation (G-KDE) to tailor for latent variables.
The G-KDE clustering algorithm learns the cluster number $K$ and separates the child-Gaussian features into distinct clusters, effectively auto-clustering hidden information within each latent dimension. 
This process can be considered data-mining within each latent dimension.
In Stage 2, using the clusters, 
we then construct the GM Probability Density Function (PDF) of the latent variable without assuming a fixed number of child distributions beforehand. 
In Stage 3, we apply the Probability Integral Transform (PIT)~\cite{davidProbabilityIntegralTransformation1948} to convert the latent variables into a uniform distribution. 

Within our proposed UT module, 
the latent space is considered in a continuous fashion 
and the non-parametric GM distribution is estimated by the G-KDE clustering algorithm, which avoids the induction bias of representation learning.
The UT module densifies the latent representation, effectively mitigating posterior collapse on a dimension-wise basis. 
This results in the decoder receiving more robust latent samples, enhancing its ability to learn and reconstruct the data. 
Importantly, the UT module operates as a self-mining process that denoises the latent representation without embedding any external inference.
Our experimental results showed that this new latent representation enhances disentanglement compared to baseline models.

In summary, the contributions of this paper are as follows:
In addressing the inefficiency of learned latent representations in VAEs, caused by irregular posterior distributions and invalid sampling of periodic data, we propose an adaptable three-stage UT module to transform irregular latent distributions into a uniform distribution to achieve better disentanglement than original latent representations from baseline models.
We have demonstrated that transforming latent variables to uniform distributions significantly improves the disentanglement of latent variables and reduces reconstruction error in VAEs.

The remainder of this paper is organized as follows: 
Section~\ref{sec:approach} formalizes the problem we addressed and details our proposed approach. 
In Section~\ref{sec:experiment}, we describe the experiments conducted, along with the results obtained, offering insights into the effectiveness of our approach. 
Finally, Section~\ref{sec:conclusion} summarizes our findings and conclusions.

\section{Proposed Approach}\label{sec:approach}
We tackle the challenge of irregular distributions within the latent space of a VAE by framing it as a constrained optimization problem.
Our proposed solution employs the G-KDE clustering algorithm to approximate the posterior GM distribution of latent variables. 
Subsequently, we leverage the PIT to facilitate the transition of these irregularly distributed latent variables into uniform distributions, 
thus enhancing the interpretability and usability of the latent space.

\subsection{Analysis of Irregular Distributions}\label{sec:problem}
\begin{figure}[htb]
    \centering
    \includegraphics[width=0.9\columnwidth]{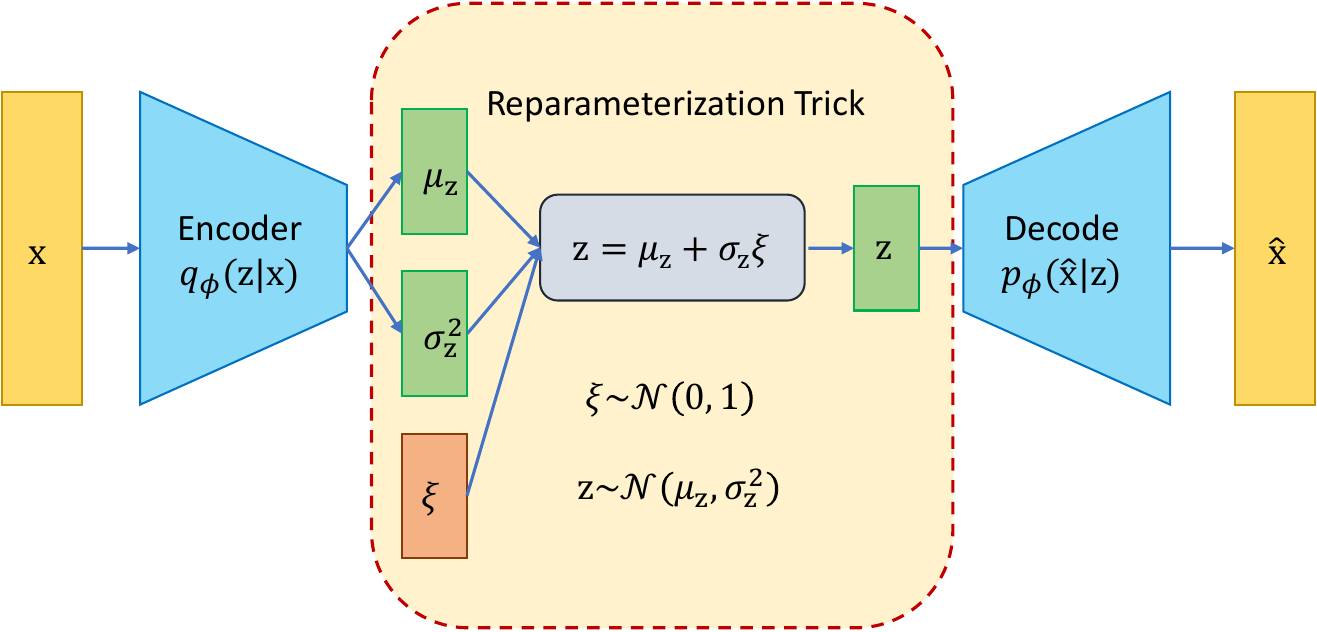}
    \caption{VAE Structure. The reparameterization trick separates the encoder output into mean $\mu_{\rz}$ and variance $\sigma_{\rz}^2$, which are then combined with an independent random variable $\xi$ to construct the latent variable $\rz$.}
    \label{fig:VAE}
\end{figure}

The $\beta$-VAE~\cite{higginsBetaVAELearningBasic2016a} extends the loss function of the original VAE~\cite{kingmaAutoEncodingVariationalBayes2013} into a constrained optimization problem by distinguishing the reconstruction loss as the objective and the KL divergence as the constraint. However, the reparameterization trick~\cite{kingmaAutoEncodingVariationalBayes2013}, a crucial aspect of the VAE structure, was not accounted for in the constraints. The reparameterization trick allows for the back-propagation of learning mean $\mu_{\rvz}$ and variance $\sigma_{\rvz}$ onto a normal distribution $p(\rvz)$ as illustrated in Fig.~\ref{fig:VAE}.

We reformulate the loss of the standard VAE into a new constrained optimization problem, treating latent variables $\rvz$ as free variables:
\begin{align}
\max_{\rvz}&\quad \mathbb{E}_{\rvz \sim \sD} \left[ \mathbb{E}_{q_{\phi}(\rvz|\rvx)} \left[ \log p_{\theta}(\hat{\rvx}|\rvz) \right] \right], \label{eqn:recon}\\
\text{subject to } &\quad
\rvz = \mu_{\rvz} + \sigma_{\rvz} \xi ,\label{eqn:reparameter}\\
&\quad \KL(q_{\phi}(\rvz|\rvx) || p(\rvz)) \leq \varepsilon. \label{eqn:kl}
\end{align}

\Eqref{eqn:recon} represents the reconstruction loss, formulated as cross-entropy, where $\rvx$ denotes sample data, $\hat{\rvx}$ is the reconstructed data, $\sD$ is the dataset, $q_{\phi}(\rvz|\rvx)$ is the posterior PDF from the encoder with parameter $\phi$, and $p_{\theta}(\hat{\rvx}|\rvz)$ is the PDF of the marginal distribution of the decoder with parameter $\theta$.

\Eqref{eqn:reparameter} introduces a hard equality constraint for the reparameterization trick, where $\mu_{\rvz}$ and $\sigma_{\rvz}$ are outputs of the encoder, representing the mean and variance of $\rvz$, respectively. $\xi\sim\mathcal{N}(0,1)$ facilitates the reparameterization trick, ensuring each latent variable value $\rz$ follows a Gaussian distribution,
\begin{equation}
    \rz\sim\mathcal{N}(\mu_{\rz}, \sigma_{\rz}^2).
\end{equation}

\Eqref{eqn:kl} is a soft inequality constraint on the KL divergence, where $\KL(\cdot\|\cdot)$ is the KL divergence operator, $q_{\phi}(\rvz | \rvx)$ is the encoder posterior, $\phi$ is the encoder parameter, $\varepsilon$ represents the slackness of the inequality constraint, ensuring the constraint is active when $\varepsilon\geq 0$, and $p(\rvz)$ is the PDF of the ideal prior distribution encoder, designed as a normal distribution $\mathcal{N}(0,1)$. Consequently, the posterior distribution $q_{\phi}(\rvz | \rvx)$ is softly bounded to the normal distribution $\mathcal{N}(0,1)$.

Our analysis examines the primal feasibility of constraints\cite{chongKarushKuhnTuckerCondition2013}, spotlighting a learned discrete data feature $\sY$ with $K$ independent classes. Represented as $\sY = \{\ry|\ry=\ry_k, k=1,2,\ldots,K\}$, each class within $\sY$ comes with its PDFs $p(\ry)$, and $\rz^{(\sY)}$ denotes the latent variable encoding the data feature $\sY$. 
The posterior distribution, when marginalized over $\sY$, is expressed as:
\begin{equation}
\begin{aligned}
    q_{\phi}(\rz^{(\sY)} | \rx) &= \sum_{\sY} q_{\phi}(\rz^{(\sY)}, \ry | \rx)\\
    &= \sum_{\sY} q_{\phi}(\rz^{(\sY)} | \ry, \rx)p(\ry | \rx).
\end{aligned}
\end{equation}
Assuming the independence of the data feature $\ry$ from the input data $\rx$, a common assumption in generative models, we have $p(\ry | \rx) = p(\ry)$, leading to:
\begin{equation}
    q_{\phi}(\rz^{(\sY)} | \rx) = \sum_{k = 1}^{K} p(\ry_k)q_{\phi}(\rz^{(\ry_k)} | \rx),
\end{equation}
where $\rz^{(\ry_k)}$ is associated with each $\ry_k$. Thus, the posterior distribution morphs into a mixture model encompassing $q_{\phi}(\rz^{(\ry_k)} | \rx)$ for $k= 1,2,\ldots,K$.

Given the constraint in \Eqr{eqn:reparameter} that each latent variable is mapped to a Gaussian distribution, $q_{\phi}(\rz^{(\ry_k)} | \rx) = \mathcal{N}(\mu_{\rz^{(\ry_k)}}, \sigma_{\rz^{(\ry_k)}}^2)$, the posterior distribution of encoder $q_{\phi}(\rz^{(\sY)} | \rx)$ is identified as a GM distribution with $p(\ry_k)$ acting as the weight for each child Gaussian $q_{\phi}(\rz^{(\ry_k)} | \rx)$.

\begin{figure}[htb]
    \centering
    \includegraphics[width=0.9\columnwidth]{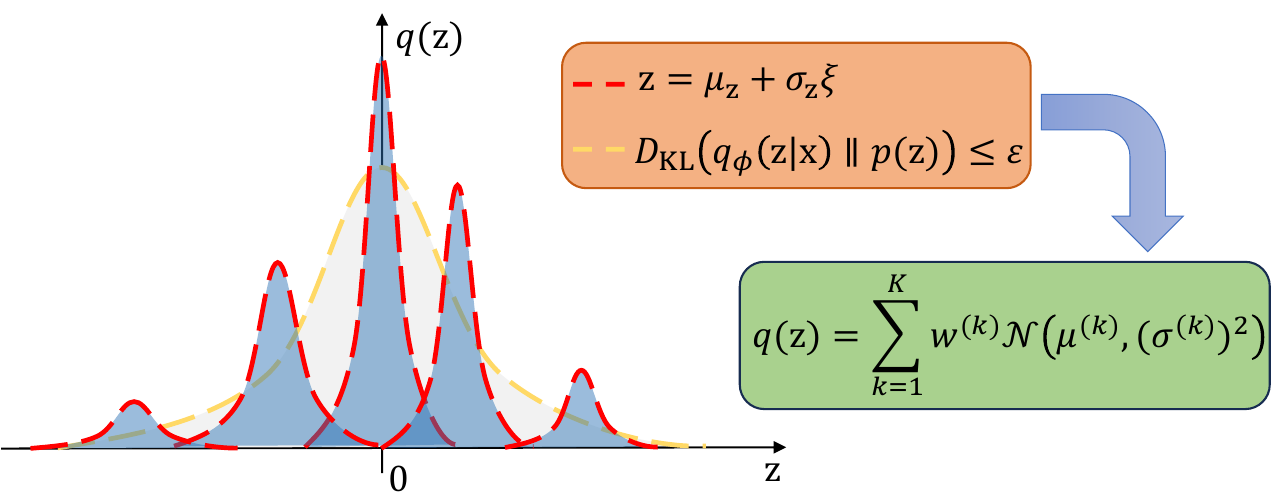}
    \caption{Illustration of GM Reasoning: 
    The yellow dashed curve represents the PDF of a normal distribution, softly constrained by \Eqr{eqn:kl}.
    The red dashed curves denote the PDFs of child Gaussian distributions, constrained by \Eqr{eqn:reparameter}. 
    The blue shadow illustrates the GM distribution of the posterior $q(\rz)$, where $w^{(k)}$, $\mu^{(k)}$, and $(\sigma^{(k)})^2$ denote the weight, mean, and variance of the $k$th child Gaussian, respectively.}
    \label{fig:GMreason}
\end{figure}

The soft constraint in \Eqr{eqn:kl} for $\rz^{(\sY)}$ is then articulated as:
\begin{equation}
    \KL{\left(\sum_{k = 1}^{K} p(\ry_k)q_{\phi}(\rz^{(\ry_k)} | \rx) \bigg\| p(\rz)\right)} \leq \varepsilon,
\end{equation}
with the aim to approximate the GM distribution to a normal distribution as illustrated in Fig.~\ref{fig:GMreason}.

Figure~\ref{fig:GMreason} elucidates the irregular distribution challenges in VAEs due to the constrained GM distributions. The encoder, by mapping discrete data features into a dense latent space, and the decoder, by expecting inputs from a normal distribution, reveal a critical discrepancy. This mismatch leads to overfitting, where the model excessively adapts to the specific characteristics of the training data, hindering generalization. 
Moreover, it fails to accurately represent periodic distributions due to the inherent assumption of Gaussian distributions, which does not align with the periodic features present in some data types.

To address these issues, our methodology aims to reconstruct GM distributions into a regular distribution, specifically opting for a uniform distribution. This approach facilitates a more generalized and flexible representation of data features, potentially enhancing the ability to capture and reproduce periodic distributions by allowing for a broader adaptation to the underlying structure of data.

\subsection{Transformation from Irregular Distribution to Uniform Distribution}
\begin{figure}[htb]
    \centering
    \includegraphics[width =0.9 \columnwidth]{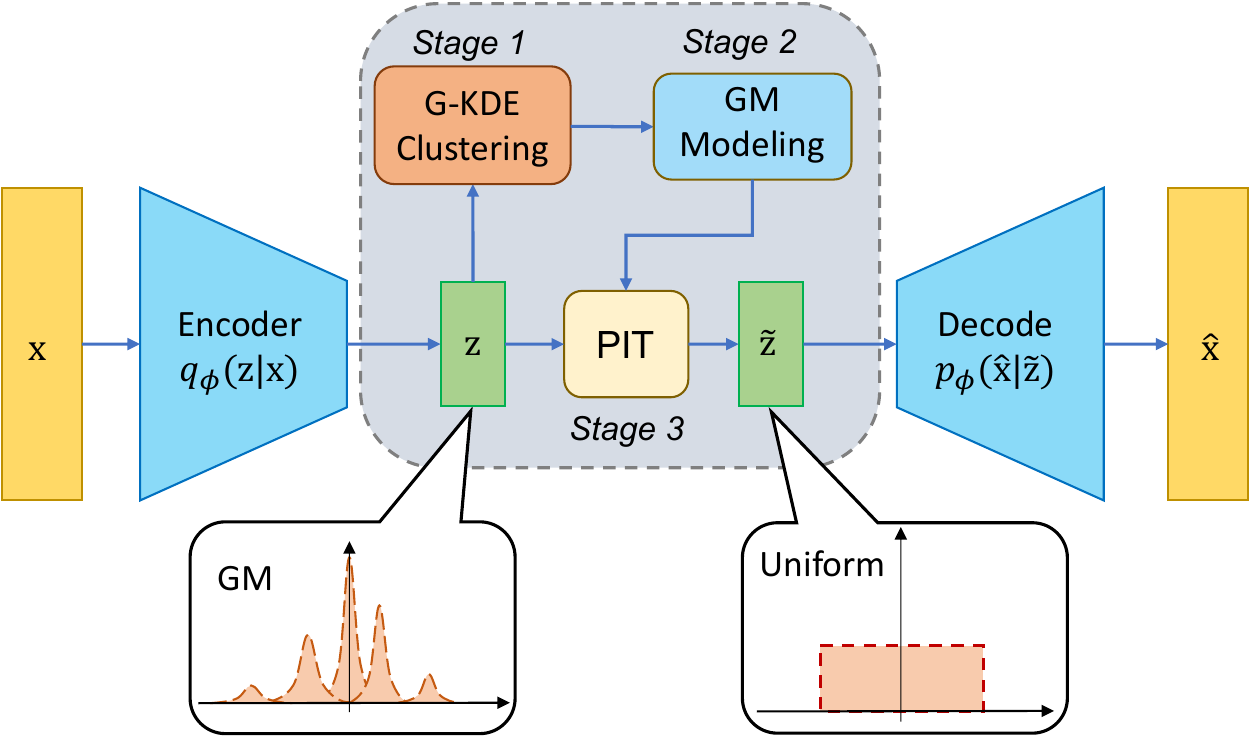}
    \caption{Schematic of the Uniform Transformation Module integrated into a VAE framework. The module, highlighted within the gray box, transforms latent variables $\rvz$ to $\tilde{\rvz}$, representing the pre- and post-transformation states, respectively.}
    \label{fig:UTModule}
\end{figure}

We present the UT module as an adaptable component designed to transform irregularly distributed latent variables to a uniform distribution. 
This module is ready to be integrated into the latent space of any VAE framework. 
The complete process highlighted in a gray box is shown in Fig.~\ref{fig:UTModule}, which illustrates the schematic of our UT Module within the VAE architecture.
It begins an input $\rx$ processed by an encoder $q_\phi(\rz|\rx)$, yielding original latent variables $\rz$. 
Our UT module operates as follows:
\begin{itemize}
    \item \textbf{Stage 1}: The latent variables $\rz$ are auto-clustered by the proposed G-KDE clustering algorithm. 
    \item \textbf{Stage 2}: A GM PDF is constructed by the estimation of obtained clusters.
    \item \textbf{Stage 3}: PIT is applied to transform $\rz$ into a uniform distribution $\tilde{\rz}$. 
\end{itemize}
Finally, the decoder $p_\phi(\rx|\tilde{\rz})$ reconstructs the output $\hat{\rx}$, completing the transformation from $\rz$ to $\tilde{\rz}$ which is a uniform distributed latent variable.

\subsection{G-KDE Clustering Algorithm and GM Modeling }

In Stage 1, we tailor the G-KDE clustering method to discern the child-Gaussian distributions within the latent-variable distribution accurately and obtain the GM model parameters.
The G-KDE clustering method, though less common in data analysis, excels at identifying irregular and intricate distributions that match GM distributions in low-dimensional and univariate cases~\cite{scaldelaiMulticlusterKDENewAlgorithm}.

G-KDE is recognized as a robust non-parametric approach for estimating the PDF of a dataset. 
Unlike parametric methods that presuppose a specific distribution shape (e.g., Gaussian), G-KDE offers the versatility to model a wide array of data patterns without adhering to predefined distributional assumptions.
Upon we identify the irregular distribution as GM, each cluster in the latent variable represents a child Gaussian.
We use the geometrical features of each cluster to obtain the parameters for each child Gaussian in the GM distribution.
Typical GM modeling methods, such as the Expectation-Maximization (EM) algorithm, require a predefined number of clusters $K$ for initialization and do not work in univariate cases such as latent variables.
The G-KDE is mathematically represented as:
\begin{equation}\label{eqn:GKDE}
\hat{f}(x) = \frac{1}{nh} \sum_{i=1}^{n} \mathcal{K}\left( \frac{x - x_i}{h} \right),
\end{equation}
where $\hat{f}(x)$ denotes the estimated PDF of $x$, $n$ symbolizes the number of data points, $x_i$ refers to the observed data points, $h$ is the bandwidth that defines the width of the Gaussian kernels, and $\mathcal{K}(\cdot)$, the Gaussian kernel function, is delineated as $\mathcal{K}(u) = \frac{1}{\sqrt{2\pi}} e^{-\frac{1}{2}u^2}$.
We use the Scott method~\cite{scottMultivariateDensityEstimation2015}, which sets the bandwidth $h=n^{-\frac{1}{5}}\sigma$, to balance smoothing and detail, ensuring an accurate representation of the underlying data distribution.

After G-KDE, we estimate the GM model parameters using the local extremas of the estimated PDF of the latent variable. 
The local maxima serve as the centroids (means) of the child Gaussians. 
The number of child Gaussians corresponds to the number of clusters, $K$.
The local minima serve as the cluster thresholds of the child Gaussians. 
The variance and weight of each child Gaussian are the variance and proportion of its respective cluster.
To detect these extremas within the G-KDE, we employ a sliding window technique.
The process of Stage 1 is succinctly captured in the following Algorithm~\ref{alg:GKDE}.

\begin{algorithm}[htb!]
\caption{G-KDE Clustering Algorithm}\label{alg:GKDE}

\begin{algorithmic}[1]
    \State \textbf{Input:} $\sZ = \{z_1, z_2, ..., z_n\}$, an array of single latent variable samples.
    \State \textbf{Output:} Cluster labels for each sample in $\mathcal{Z}$, cluster centroids $\{\hat{\mu}^{(k)}\}$, variances $\{\left(\hat{\sigma}^{(k)}\right)^2\}$, and weights $\{\hat{w}^{(k)}\}$.
    \State \textbf{Procedure:}
    \State Estimate the PDF $\hat{f}(z)$ of $\sZ$ using G-KDE as \Eqr{eqn:GKDE}:
    \[
    \hat{f}(z) = \frac{1}{nh} \sum_{i=1}^{n} \mathcal{K}\left( \frac{z - z_i}{h} \right).
    \]
     
    \State Compute the local maxima $\{\hat{\mu}^{(k)}\}$ and minima $\{\delta_j\}$ of $\hat{f}(z)$, where $j=1,2,\cdots, K-1$.

    \For{each $z_i$}
        \If{$z_i \leq \delta_1$}
            \State $z_i$ belongs to Cluster $1$.
        \EndIf
        \For{$j = 1$ to $K-1$}
            \If{$\delta_j < z_i \leq \delta_{j+1}$}
                \State $z_i$ belongs to Cluster $j+1$.
            \EndIf
        \EndFor
        \If{$z_i > \delta_{K-1}$}
            \State $z_i$ belongs to Cluster $K$.
        \EndIf
    \EndFor

    \For{each Cluster $k$}
        \State Compute variance for Cluster $k$:
        $$
        \left(\hat{\sigma}^{(k)}\right)^2 = \frac{1}{|\sZ^{(k)}|-1} \sum_{z \in \sZ^{(k)}} (z - \hat{\mu}^{(k)})^2,
        $$
        \indent where $|\sZ^{(k)}|$ is the number of samples in Cluster~$k$.
        \State Compute weight for Cluster $k$:
        $$
        \hat{w}^{(k)} = \frac{|\sZ^{(k)}|}{n}.
        $$
    \EndFor

\end{algorithmic}
\end{algorithm}

Utilizing Algorithm~\ref{alg:GKDE}, we acquire all parameters of the GM distribution and each PDF of child Gaussian is $\hat{q}_{k}\left(\rz\right) = \mathcal{N}\left(\hat{\mu}^{(k)}, \left(\hat{\sigma}^{(k)}\right)^2\right) $.
In Stage 2, we subsequently model and reconstruct the posterior GM distribution as:
\begin{equation}\label{eqn:GMPDF}
\hat{q}_{\phi}(\rz| \rx) = \sum_{k=1}^{K} \hat{w}^{(k)} 
\hat{q}_{k}\left(\rz\right).
\end{equation}
Following this reconstruction, the subsequent stage involves transforming this GM distribution into a uniform distribution. 

\subsection{PIT of Latent Variable}
In Stage 3, the transformation of the latent variable employs the PIT~\cite{davidProbabilityIntegralTransformation1948} -- a statistical technique that normalizes any given random variable to a uniform distribution across the interval $[0, 1]$. This method is fundamentally based on the properties of the integral of monotonically increasing functions and is extensively utilized in the domain of probability density research~\cite{papamakariosNormalizingFlowsProbabilistic2021}.

The PIT is predicated on the concept that a random variable $x$, associated with a CDF $F(x)$, will be transformed such that the resulting variable $\tilde{x}$ adheres to a uniform distribution, succinctly expressed as $\tilde{x} = F(x) \sim \mathcal{U}(0, 1)$.

Following this principle, the transformed latent variable is given by
\begin{equation}
    \label{eqn:uniformZ}
    \tilde{\rz} =\sum_{k=1}^{K} \hat{w}^{(k)} \hat{Q}_{k}\left(\rz\right),
\end{equation}
where $\hat{Q}_{k}\left(\rz\right)$ represents the CDF of the child Gaussian $\hat{q}_{k}\left(\rz\right)$ corresponding to each cluster $k$. This equation effectively maps the latent variable $\rz$ through the CDFs of its Gaussian components, weighted by their respective probabilities $\hat{w}^{(k)}$, to produce a uniformly distributed transformation $\tilde{\rz}$.
In our implementation, we normalize $\tilde{\rz}$ to the range $[-4, 4]$ to prevent gradient vanishing at the zero point across various activation functions.

\section{Experimental Results}\label{sec:experiment}
\subsection{Disentanglement}
To validate our proposed UT module, we performed computer simulations to focus on evaluating the learned representations from our proposed UT model against several baseline VAE models, utilizing the dSprites dataset~\cite{higginsBetaVAELearningBasic2016a} for disentanglement testing, a widely recognized benchmark in the field of disentanglement studies.
To extend our analysis beyond synthetic data, we incorporated the Modified National Institute of Standards and Technology (MNIST) dataset~\cite{lecun2010mnist}, a staple in machine learning research. 
Utilizing labels of MNIST, we conducted an examination of disentanglement in a real-world context, offering a comprehensive view of the performance of our proposed module in a VAE across both synthetic and non-synthetic datasets.

The baseline models selected for comparison include the Vanilla VAE, $\beta$-VAE, FactorVAE, and TCVAE.
Each of these models represents a major approach to encouraging disentanglement in the latent space.
And TCVAE is recognized as the state-of-the-art model in VAE disentanglement.
The configuration details are provided in Appendix~\ref{apx:experiment}.
To quantify the disentanglement performance of each model, 
we employed most popular metrics, including:
Mutual Information Gap (MIG)~\cite{chenIsolatingSourcesDisentanglement2019}, 
$\beta$-VAE Metric~\cite{higginsBetaVAELearningBasic2016a},
FactorVAE Metric~\cite{kimDisentanglingFactorising2018}, 
Total Correlation (TC)~\cite{kimDisentanglingFactorising2018}, and Total Maximum Correlation (TMC)\cite{nguyenMultivariateMaximalCorrelation2014}.

The results delineated in Table~\ref{tab:model_comparison_dsprites} critically examine the efficacy of the proposed UT module in augmenting disentangled representations.
These representations ostensibly encapsulate the intrinsic generative factors of the dataset, thus signifying a progression from the baseline models. 
The incorporation of the UT module across various VAE frameworks predominantly enhances metric performance, signifying an amplification in disentanglement and data delineation.

\begin{table}[ht]
\centering
\caption{Disentanglement Performance on dSprites}
\label{tab:model_comparison_dsprites}
\begin{tabular}{lccccc}
\hline
\textbf{Model} & \textbf{MIG}& \textbf{$\beta$-VAE}& \textbf{FactorVAE}& \textbf{TC}&\textbf{TMC}\\
\hline
VAE& 0.23& 0.76& 0.77& 3.68&2.77\\ 
VAE+UT& 0.23& \textbf{0.90} $\uparrow$& 0.79 $\uparrow$& 3.69 $\uparrow$&2.97 $\uparrow$\\
\hline
$\beta$-VAE& 0.42& 0.62& 0.77& 3.74&3.38\\
$\beta$-VAE+ UT& 0.43 $\uparrow$& 0.68 $\uparrow$& 0.77& \textbf{3.77} $\uparrow$&\textbf{3.42} $\uparrow$\\
\hline
FactorVAE& 0.48& 0.38& \textbf{0.87}& 3.12&2.92\\
FactorVAE+UT& \textbf{0.49}$\uparrow$& 0.36 $\downarrow$& 0.86 $\downarrow$& 3.14 $\uparrow$&2.98 $\uparrow$\\
\hline
TCVAE& 0.36& 0.37                 & \textbf{0.87}                  & 3.16&2.95\\
TCVAE+UT& 0.36& 0.46 $\uparrow$                 & 0.84 $\downarrow$                 & 3.20 $\uparrow$&3.00 $\uparrow$\\
\hline
\end{tabular}
\end{table}

Notwithstanding, the UT module impact on the performance metrics of foundational VAE models exhibits discernible complexity. For instance, the $\beta$-VAE model demonstrates substantial improvements upon UT module integration, as evidenced by upward trends ($\uparrow$). Conversely, FactorVAE and TCVAE models display a decrement in specific metrics ($\downarrow$), highlighting the non-uniform efficacy of the UT module. Such disparities may be attributed to the inherent design of FactorVAE and TCVAE, which intrinsically regulate latent variables into a balanced GM through discriminators and bespoke loss functions, potentially rendering them less receptive to subsequent transformations via the UT module. 

The further analysis reveals that latent representations in VAE and $\beta$-VAE models remain robustly competitive post-UT transformation, even when contrasted with contemporaneous models.
The findings align with recent comparative studies~\cite{chadebecPythaeUnifyingGenerative2023} indicating that merely altering the encoder and decoder configurations within VAE-based frameworks does not necessarily translate to improved generative capabilities. 

\begin{figure}[htbp!]
    \centering
    \subfloat[VAE]{
        \includegraphics[width=0.45\columnwidth]{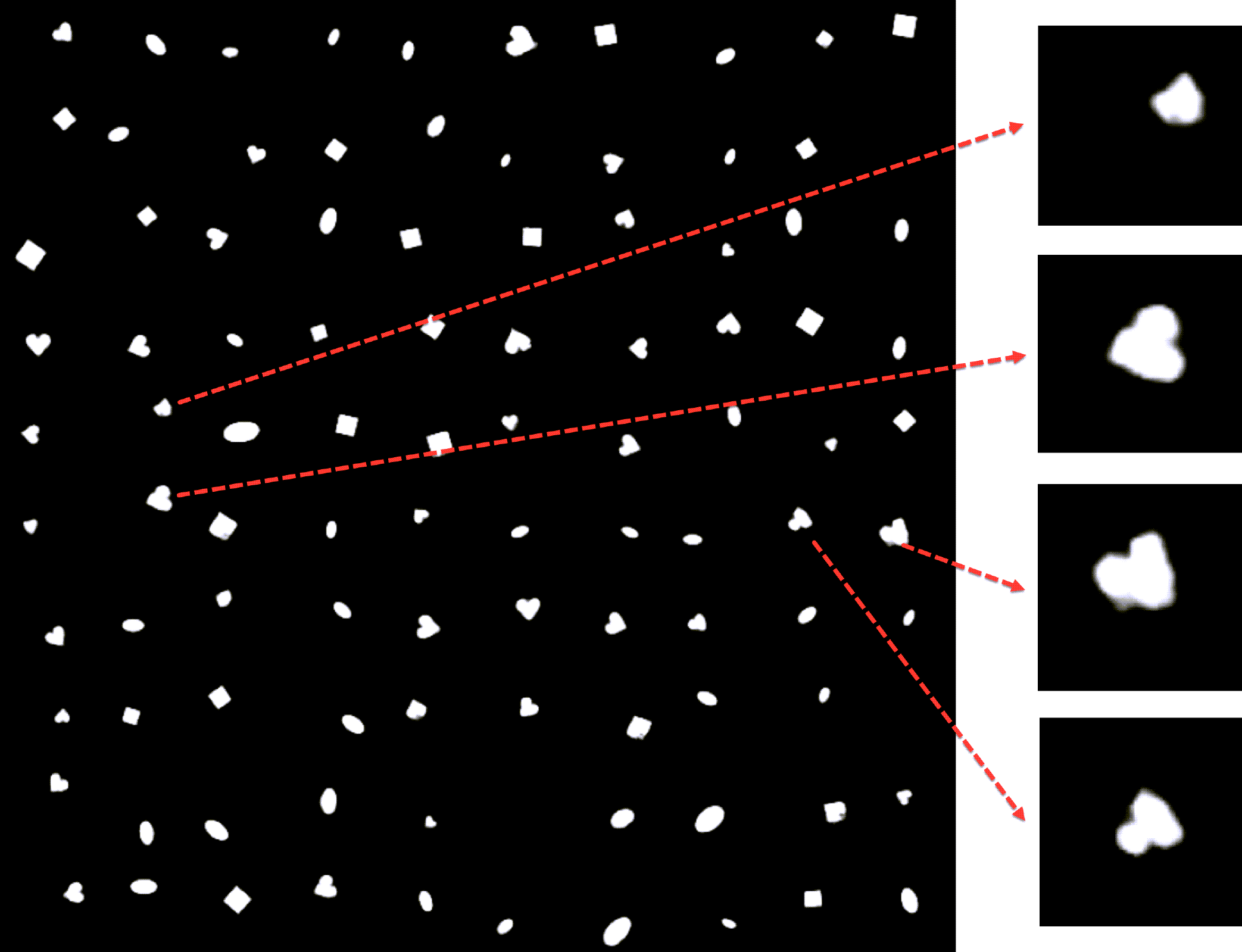}
        \label{fig:vae}
    }
    \subfloat[VAE+UT]{
        \includegraphics[width=0.45\columnwidth]{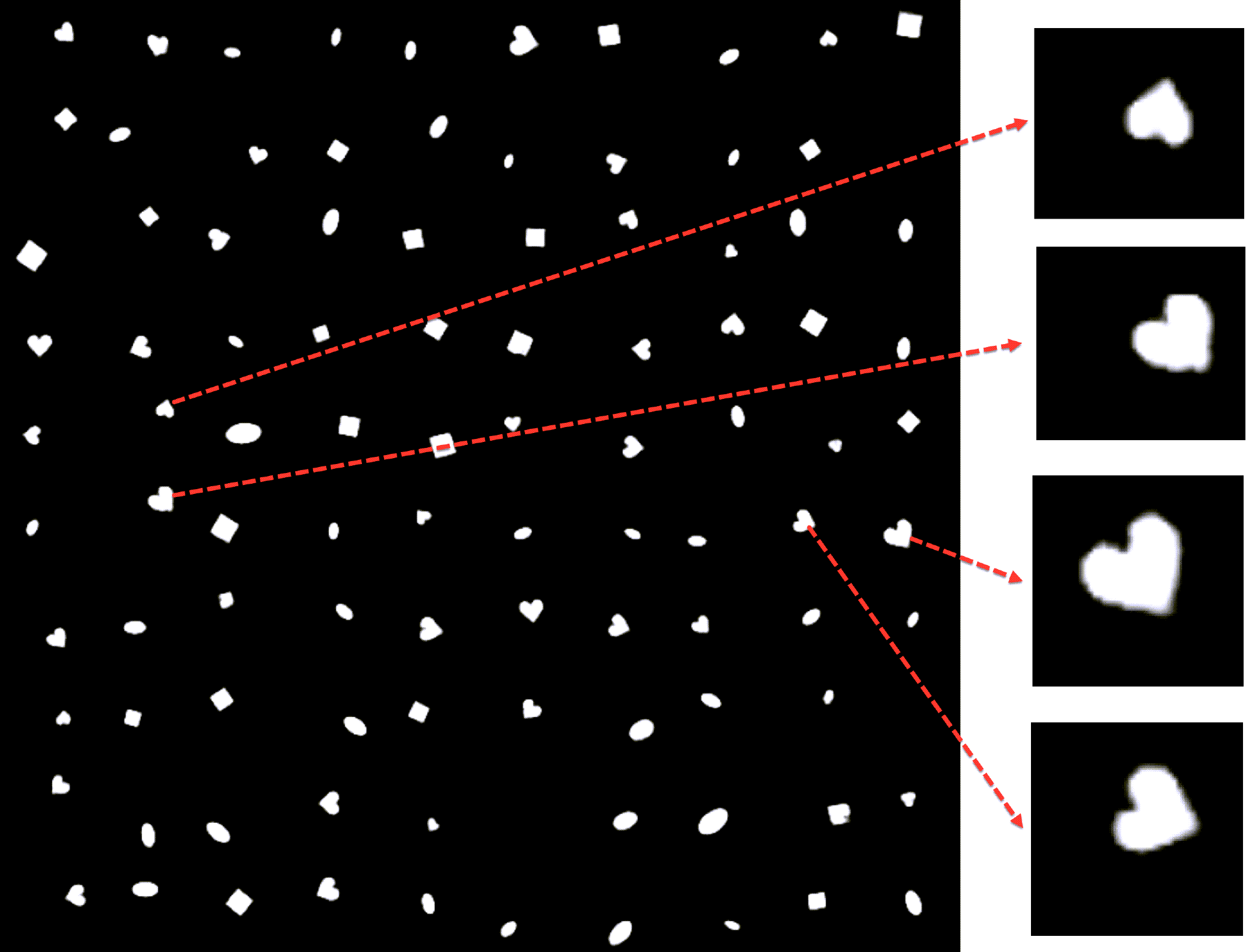}
        \label{fig:vaeut}
    }
    \caption{Comparative Generative Results on dSprites Dataset using VAE Models with a Fixed Random Seed.}
    \label{fig:vaegenerative}
\end{figure}

We further conducted a qualitative analysis comparing the reconstruction quality of the VAE and $\beta$-VAE models, both prior to and following the application of the UT module. Notably, the average reconstruction loss decreased from 29.31 to 27.43 in the VAE model and from 36.06 to 34.28 in the $\beta$-VAE model, indicating enhanced performance post-UT application. 
A qualitative comparison of the generative capabilities is illustrated in Fig.~\ref{fig:vaegenerative}, showcasing samples generated by each model. 
The VAE+UT model demonstrates superior performance in reconstructing heart-shaped objects within the dSprites dataset.
The application of the UT module addresses the issue of invalid solutions when samples are underrepresented, leading to sharper reconstructions on edge samples by the decoder.

Table~\ref{tab:model_rotation_scores_dsprites} examines the impact of the UT module on the rotation feature in the dSprites dataset, demonstrating that the UT module improves the encoding of periodic features.

\begin{table}[ht]
\centering
\caption{Disentanglement Scores of Rotation Feature of dSprites}
\label{tab:model_rotation_scores_dsprites}
\begin{tabular}{lcccc}
\hline
\textbf{Metric}& \textbf{$\beta$-VAE}& \textbf{FactorVAE}& \textbf{TC}&\textbf{TMC}\\
\hline
VAE& 0.40& 0.58& 0.28&0.14\\ 
VAE+UT& \textbf{1.0} $\uparrow$& 0.58& \textbf{0.32}$\uparrow$&\textbf{0.17}$\uparrow$\\
\hline
$\beta$-VAE& 0.30& 0.53& 0.31&0.14\\
$\beta$-VAE+ UT& \textbf{1.0} $\uparrow$& 0.55$\uparrow$& 0.31&0.16$\uparrow$\\
\hline
FactorVAE& 0.09& \textbf{0.70}& 0.15&0.11\\
FactorVAE+UT& 0.11$\uparrow$& 0.53$\downarrow$& 0.18$\uparrow$&0.13$\uparrow$\\
\hline
TCVAE& 0.20& \textbf{0.70}& 0.12&0.09\\
TCVAE+UT& 0.40$\uparrow$& 0.65$\downarrow$& 0.13$\uparrow$&0.11$\uparrow$\\
\hline
\end{tabular}
\end{table}

Table~\ref{tab:model_comparison_MNIST} shows that the UT module affects disentanglement metrics in VAE models on the MNIST dataset, which lacks labeled primitive factors unlike dSprites.
Hence, MNIST disentanglement scores reflect the correspondence between labeled features and latent variables. 
UT generally elevates the MIG scores, especially in the $\beta$-VAE+UT model, suggesting enhanced disentanglement. 
However, $\beta$-VAE and FactorVAE metrics are inferred from data labels, reflecting potential latent variable relevance. 
TCVAE maintains consistent TC and TMC performance, showing its robustness and unaffected disentanglement by UT integration.

\begin{table}[ht]
\centering
\caption{Disentanglement Performance on MNIST}
\label{tab:model_comparison_MNIST}
\begin{tabular}{lccccc}
\hline
\textbf{Model} & \textbf{MIG}& \textbf{$\beta$-VAE}& \textbf{FactorVAE}& \textbf{TC}&\textbf{TMC}\\
\hline
VAE& 1.1e-3& 0.60& 0.20& 0.96&0.086\\
VAE+UT& 1.8e-3$\uparrow$& 0.60 $\uparrow$& 0.20& 0.95$\downarrow$&0.085 $\downarrow$\\
\hline
$\beta$-VAE& 1.3e-3& 0.40& 0.20& 0.96&0.081\\
$\beta$-VAE+ UT& \textbf{2.4e-3} $\uparrow$& 0.60 $\uparrow$& 0.20& 0.94$\downarrow$&0.080$\downarrow$\\
\hline
FactorVAE& 1.5e-3& 0.60& 0.20& 0.96&0.087\\
FactorVAE+UT& 2.2e-3$\uparrow$& 0.40 $\downarrow$& 0.20& 0.95$\downarrow$&0.088$\uparrow$\\
\hline
TCVAE& 1.9e-3& 0.60& 0.20& \textbf{1.0}&\textbf{0.11}\\
TCVAE+UT& 2.3e-3$\uparrow$& 0.40 $\downarrow$& 0.20& \textbf{1.0}&\textbf{0.11}\\
\hline
\end{tabular}
\end{table}

Our UT module runs on a CPU, taking approximately 106 seconds for dSprites and 37 seconds for MNIST to process the entire training set. This is reasonably fast, with a runtime comparable to training an epoch on a GPU.

\subsection{Clustering and Distribution}
\begin{figure}[htbp!]
    \centering
    \subfloat[Histograms of the Clustered Original Distribution.]{
        \includegraphics[width=0.9\columnwidth]{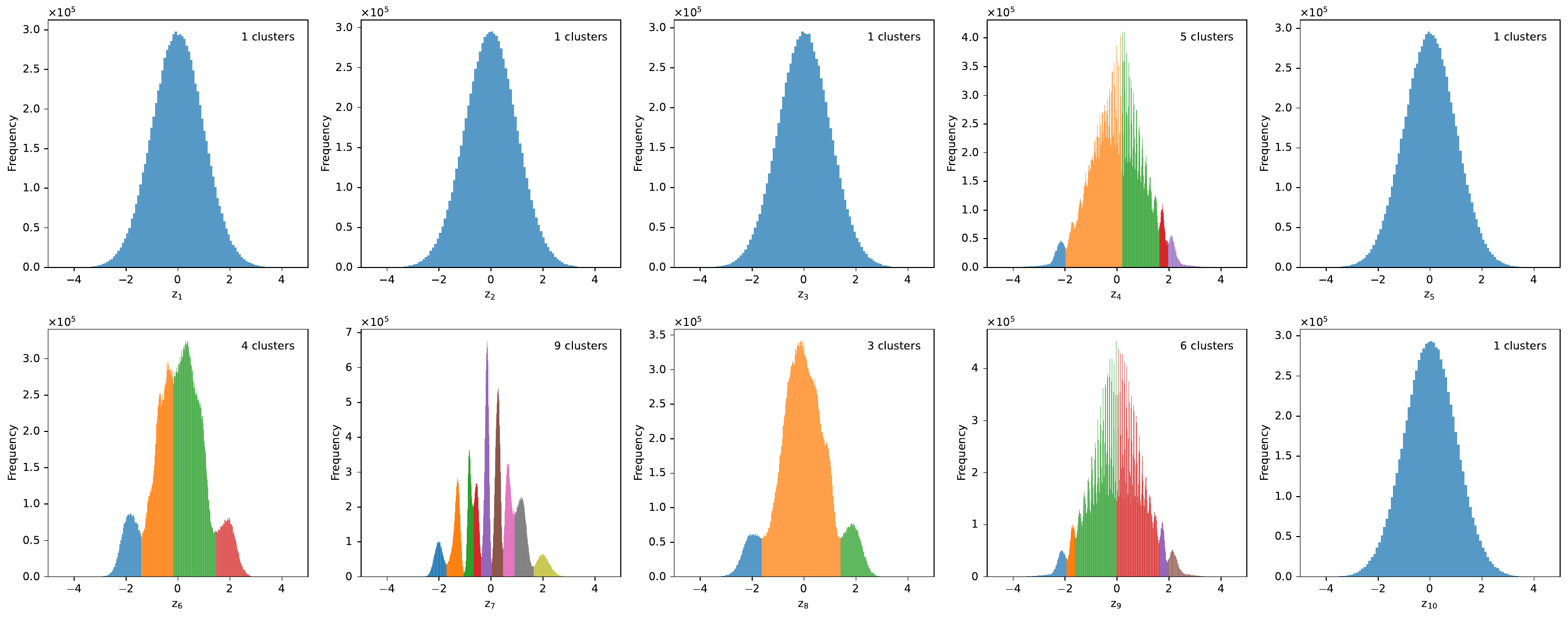}
        \label{fig:zcluster}
    }
    
    \subfloat[Histograms after the Uniform Transformation]{
        \includegraphics[width=0.9\columnwidth]{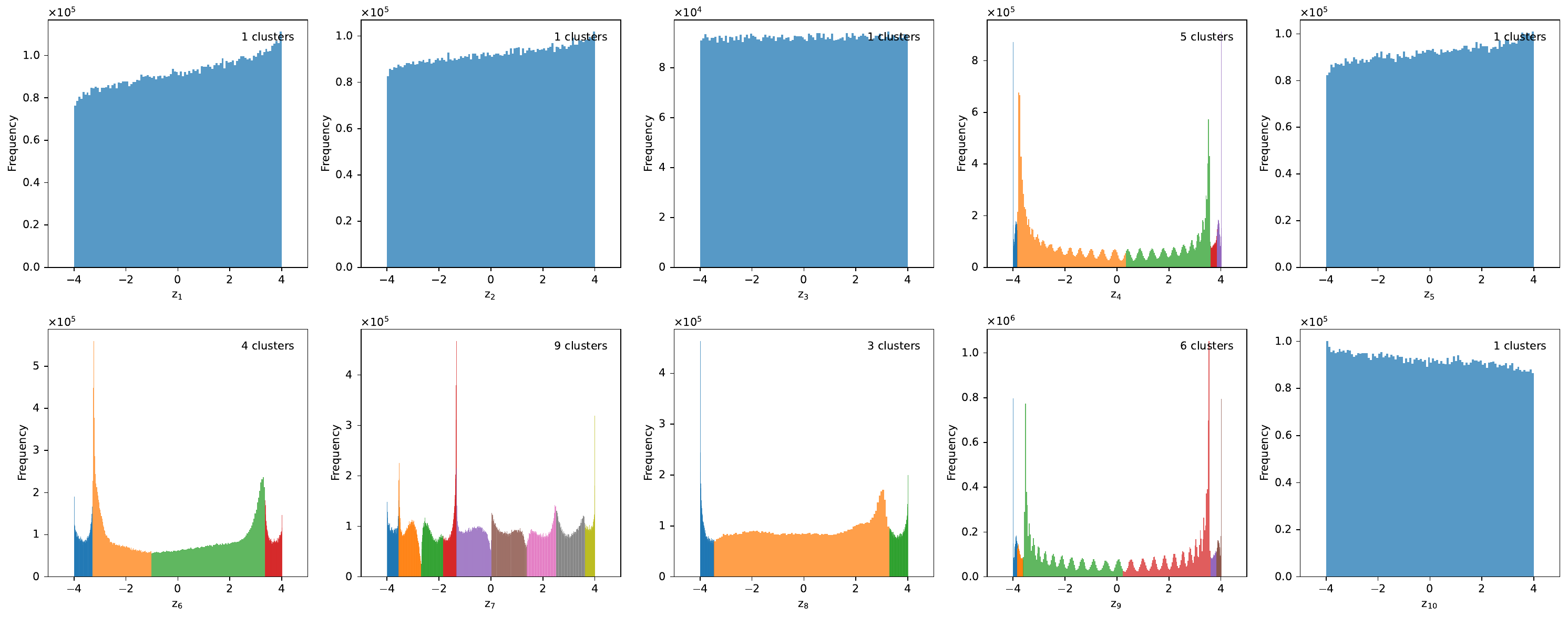}
        \label{fig:utzcluster}
    }
    
    \caption{Clusters of Latent Variables of the Replication of $\beta$-VAE of dSprites Dataset. 
    Clusters are separated by different colors in each histogram.
    The number of clusters are given at the top right corner of each histogram.
    }
    \label{fig:clusters}
\end{figure}

We specifically examined the impact of our G-KDE clustering algorithm on the latent variable distributions of $\beta$-VAE model.
Initially, the latent variable distribution, as depicted in the original histogram (Fig.~\ref{fig:betaVAE_dist}) provide a baseline understanding of the latent representation space. 
Our findings, illustrated in Figures \ref{fig:zcluster} and \ref{fig:utzcluster}, highlight the transformation effects before and after applying the G-KDE clustering and the UT module.
Figure \ref{fig:zcluster} displays the clustered histograms of the original latent variable distributions from the $\beta$-VAE model. 
Figure \ref{fig:utzcluster} reveals the latent variable distributions following the application of our G-KDE clustering algorithm and subsequent uniform transformation. 

Figure~\ref{fig:utzcluster} shows a noticeable sharp increase at the edges of each transformed child distribution. 
This sharp increase is attributed to the error in variance estimation. To address this issue, an ad-hoc solution could involve employing a sigmoid function at the PIT module to smoothen the edges. 
However, it is important to note that this error did not impact our experimental results.
Additionally, the clustering results reveal posterior collapse instances, characterized by singular-count clusters, implying a lack of encoded information in certain latent dimensions.

After the G-KDE clustering and PIT, the latent-variable distribution becomes more uniform, enhancing the interpretability and suggesting an improvement in the disentanglement capabilities.
This indicates that the UT module substantially bolsters the VAE ability to more accurately encapsulate and articulate the underlying data features.

\section{Conclusion}\label{sec:conclusion}
This paper introduces an adaptable UT module integrating G-KDE clustering, GM modeling, and PIT within the VAE architecture to enhance latent representation. 
The non-parametric G-KDE algorithm is crucial for identifying and structuring the latent space by clustering encoded variables.
This sets the stage for GM modeling, reflecting the intrinsic data distribution. 
Applying PIT to these clusters transforms the GM distribution into a uniform one, resolving irregular distributions and ensuring each latent dimension contributes effectively to data generation.
The empirical results underscore the utility of the UT module in producing a more structured latent space that can capture the generative factors of the data with greater precision.

Subsequent work will aim to broaden the scope of the UT module, applying it to datasets of increased complexity and a wider range of representation learning applications. 

\bibliographystyle{IEEEtran}

\bibliography{mybibfile}
\appendices

\section{$\beta$-VAE Replication}\label{apx:betavae}
\begin{figure}[htb]
    \centering
    \includegraphics[width=0.8\columnwidth]{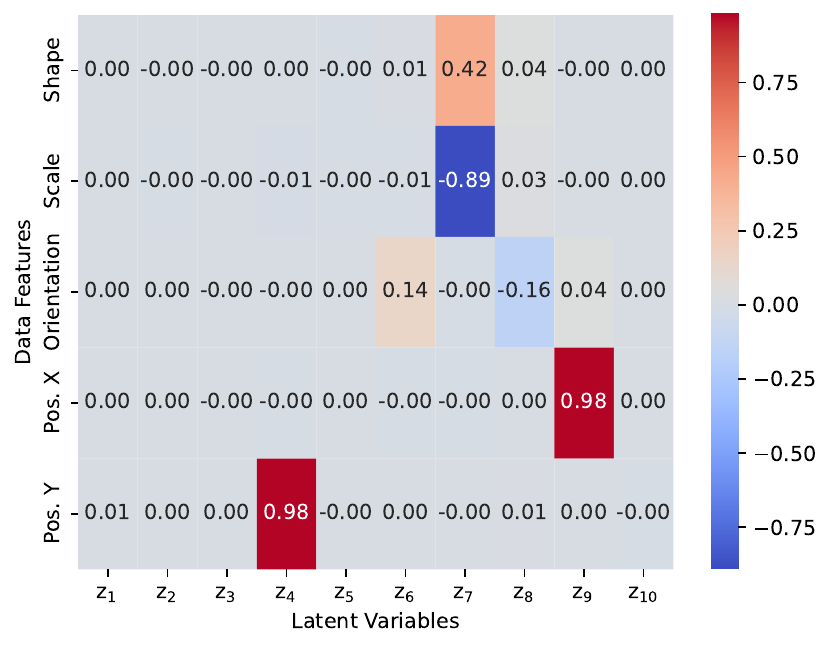}
    \caption{Heatmap of Correlation Between Latent Variables and Data Features.}
    \label{fig:betacorrelation}
\end{figure}

We meticulously replicated the $\beta$-VAE model of dSprites dateset as described in the original $\beta$-VAE paper~\cite{higginsBetaVAELearningBasic2016a}, with a latent dimensionality of $10$ and the KKT parameter $\beta$ set to $4$. The heatmap illustrating the correlation between the latent variables and data features is shown in Fig.~\ref{fig:betacorrelation}. The latent variable represents the data feature with the highest correlation value. 
The histograms of the latent variables are provided in Fig.~\ref{fig:betaVAE_dist}.

\section{Experiment Setup}\label{apx:experiment}
We used a symmetric VAE architecture for all models to ensure consistency. 
The architecture details for dSprites and MNIST datasets are provided in Table \ref{tbl:settings}.
Model-specific loss parameters were adopted from original publications without fine-tuning:
\begin{itemize}
    \item $\beta$-VAE: $\beta=4$
    \item FactorVAE: $\gamma = 10$
    \item TCVAE: $\beta = 2$, $\alpha = 1$, and $\gamma = 1$
\end{itemize}

\begin{table}[htb!]
    \centering
    \caption{Autoencoder Configuration for dSprites and MNIST}
    \label{tbl:settings}
    \begin{tabular}{lp{3cm}p{3cm}}
        \hline
        \textbf{Aspect} & \textbf{dSprites} & \textbf{MNIST} \\
        \hline
        Dataset & dSprites & MNIST \\
        Optimizer & Adam, LR: 1e-3 & Adam, LR: 1e-4 \\
        Batch Size & 64 & 64 \\
        Max Epochs & 100 (early stopping) & 300 (early stopping) \\
        \hline
        Input & $64\times 64$ grayscale & $28\times 28$ grayscale \\
        Encoder & 4 Conv layers $32\times 4\times 4$, FC 784 to 400, ReLU & 4 Conv layers $128\times 4\times 4$, BN after each Conv, FC 2048 to 1024, ReLU \\
        Latent Space & 10 dimensions & 32 dimensions \\
        Decoder & Deconv layers mirror encoder, ReLU & Deconv layers mirror encoder, ReLU \\
        Output & $64\times 64$ grayscale, Sigmoid & $28\times 28$ grayscale, Sigmoid \\
        \hline
    \end{tabular}
\end{table}
The code of UT module is available at \url{github.com/supershiye/UniformTransformationModule}.

\addtolength{\textheight}{-12cm}   




\end{document}

%% file: math_commands.tex
\usepackage{amsmath,amsfonts,bm}

\def\Figr#1{Fig.~\ref{#1}}

\def\Eqref#1{Equation~(\ref{#1})}
\def\Eqr#1{Eq.~\eqref{#1}}

\def\1{\bm{1}}

\def\rx{{\textnormal{x}}}
\def\ry{{\textnormal{y}}}
\def\rz{{\textnormal{z}}}

\def\rvx{{\mathbf{x}}}

\def\rvz{{\mathbf{z}}}

\DeclareMathAlphabet{\mathsfit}{\encodingdefault}{\sfdefault}{m}{sl}
\SetMathAlphabet{\mathsfit}{bold}{\encodingdefault}{\sfdefault}{bx}{n}

\def\sD{{\mathbb{D}}}

\def\sY{{\mathbb{Y}}}
\def\sZ{{\mathbb{Z}}}

\newcommand{\KL}{D_{\mathrm{KL}}}

